\documentclass[letterpaper, 10 pt, conference]{ieeeconf}
\IEEEoverridecommandlockouts
\title{Reinforcement Learning for Legged Robots:\\Motion Imitation from Model-Based Optimal Control}
\author{AJ Miller$^\dagger$, Shamel Fahmi$^\dagger$, Matthew Chignoli, and Sangbae Kim$^*$%
\thanks{$^\dagger$Equal contribution.
$^*$The authors are with the Biomimetic Robotics Lab, Massachusetts Institute of Technology (MIT), Cambridge, MA, USA. 
(email: \{sfahmi, ajm4, chignoli\}@mit.edu).}}
\usepackage[final]{hyperref} 
\hypersetup{colorlinks=true, citecolor=MidnightBlue, linkcolor=MidnightBlue, urlcolor=MidnightBlue}
\usepackage{cite} 
\usepackage{graphicx} 
\graphicspath{{./includes/}, {./includes/supplementary_includes/}}
\usepackage[]{units} 
\usepackage{xspace} 
\usepackage{amsfonts} 
\usepackage[usenames,dvipsnames]{xcolor} 
\usepackage{tabularx}
\usepackage{multicol}
\usepackage[nonumberlist, nogroupskip, section=section, numberedsection=autolabel]{glossaries} 
\newacronym{to}{TO}{Trajectory Optimization}
\newacronym{rl}{RL}{Reinforcement Learning}
\newacronym{wbc}{WBC}{Whole-Body Control}
\newacronym{mpc}{MPC}{Model Predictive Control}
\newacronym{cmpc}{cMPC}{convex MPC}
\newacronym{lmpc}{lMPC}{linear MPC}
\newacronym{mocap}{MoCap}{Motion Capture}
\newacronym{pmtg}{PMTG}{Policies Modulating Trajectory Generator}
\newacronym{cpg}{CPG}{Central Pattern Generator}
\newacronym{grf}{GRF}{Ground Reaction Force}
\newacronym{mimoc}{MIMOC}{Motion Imitation from Model-Based Optimal Control}
\newacronym{ac}{AC}{Actor-Critic}
\newacronym{ppo}{PPO}{Proximal Policy Optimization}
\newacronym{mlp}{MLP}{Multi-Layer Perceptron}
\newacronym{mdp}{MDP}{Markov Decision Process}
\newacronym{dofs}{DoFs}{Degrees of Freedom}
\makeglossaries
\newcommand{\ig}{IsaacGym\xspace}
\newcommand{\mc}{Mini-Cheetah\xspace}
\newcommand{\humanoid}{MIT Humanoid\xspace}
\newcommand{\RS}{Cheetah-Software\xspace}
\newcommand{\ie}{i.e.,\xspace}
\newcommand{\Rnum}{\mathbb{R}} 
\newcommand{\eref}[1]{(\ref{#1})} 
\newcommand{\fref}[1]{Fig.~\ref{#1}} 
\newcommand{\vref}[1]{Video~{#1}} 
\newcommand{\tref}[1]{Table~\ref{#1}} 
\newcommand{\boldSubSec}[1]{\noindent\textbf{#1.}}

\begin{document}
\maketitle

\begin{abstract}
We propose \acrshort{mimoc}: Motion Imitation from Model-Based Optimal Control. 
\acrshort{mimoc} is a \acrfull{rl} controller
that learns agile locomotion by imitating reference trajectories from model-based optimal control. 
\acrshort{mimoc} mitigates challenges faced by other motion imitation \acrshort{rl} approaches
because the references are dynamically consistent, require no motion retargeting, and include torque references. 
Hence, \acrshort{mimoc} does not require fine-tuning.
\acrshort{mimoc} is also less sensitive to modeling and state estimation inaccuracies than model-based controllers.
We validate~\acrshort{mimoc} 
on the \mc in outdoor environments over a wide variety of challenging terrain, and on the \humanoid in simulation.
We show 
cases where \acrshort{mimoc} outperforms model-based optimal controllers,
and show that imitating torque references improves the policy's performance.
\end{abstract}

\section{Introduction}
Legged robots have shown remarkable agile capabilities
in academia~\cite{Lee2021,Yang2020,Semini2019,Katz2019,Bledt2018,Hutter2016}
and in industry~\cite{BostonDynamics2021,AgilityRobotics2021,UniTree2021}.
Locomotion control for legged robots usually relies on model-based optimal control approaches
such as~\gls{wbc}~\cite{Kim2019, Fahmi2019, Kuindersma2014},
\gls{mpc}~\cite{Sleiman2021, Bledt2020, DiCarlo2018}, 
and~\gls{to}~\cite{Melon2021, Ponton2021, Mastalli2020b, Winkler2018b}.
Recently,~\gls{rl} for locomotion control has proven to be robust 
and reliable~\cite{Rudin2022, Yu2022, Miki2021, Gangapurwala2021, Kumar2021, Tsounis2020}.
To date, no dominant approach among model-based,~\gls{rl}, or hybrid methods has emerged in the field of legged robots.

Model-based approaches originate from physics and are interpretable, 
allowing a skilled engineer to directly evaluate and debug controllers. 
\gls{rl} approaches typically lack a model and are more challenging to interpret, design, and tune.
However, they are also less sensitive to model inaccuracies.
A major advantage of~\gls{rl} is that the agent learns from its mistakes during training.
Simulated noisy observations
and randomized physical parameters of the robot and the environment
teach the policy to overcome modeling and state estimation inaccuracies better than model-based approaches.

\begin{figure}\centering
\includegraphics[width=\columnwidth]{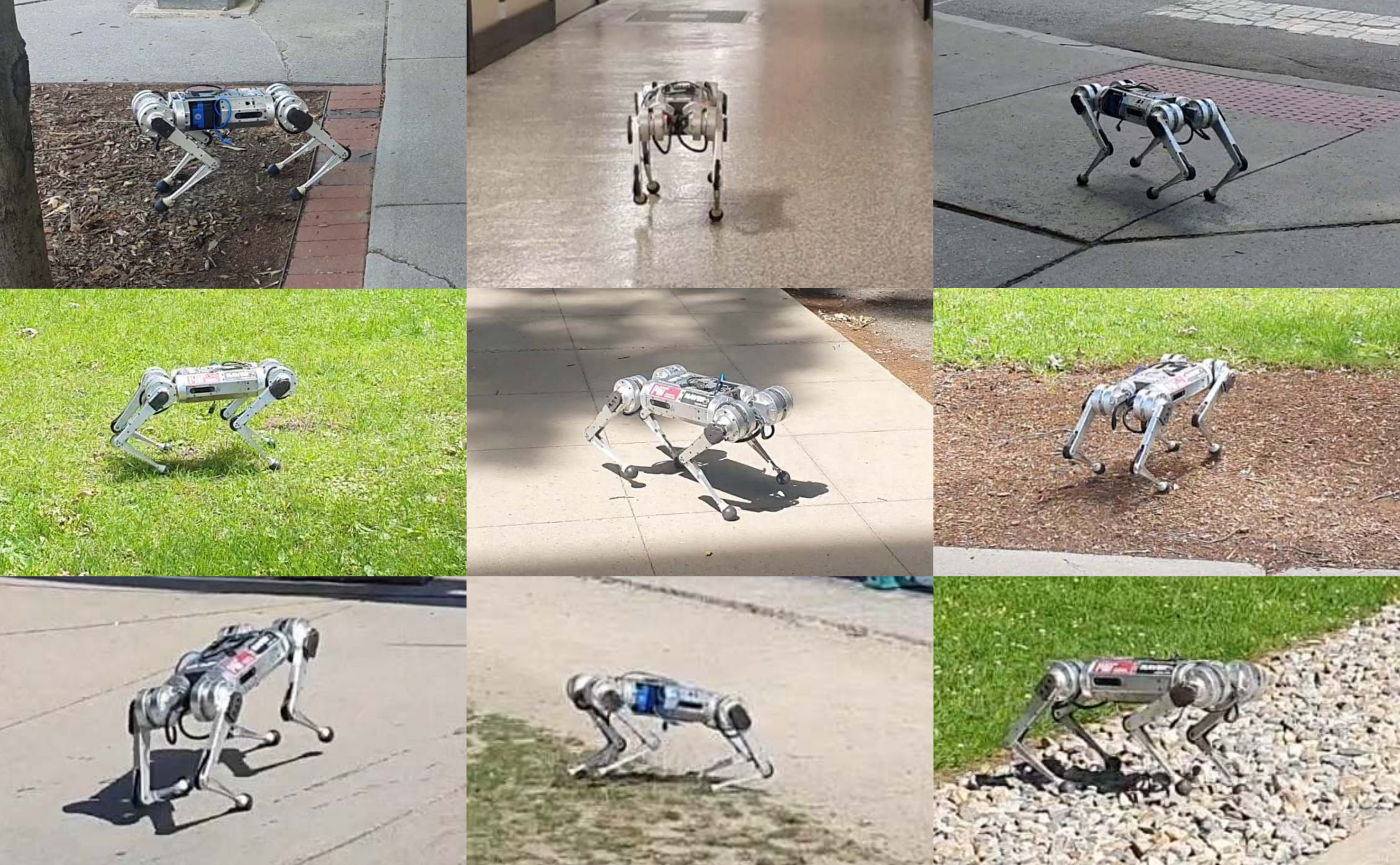}		
\caption{This figure shows multiple screenshots of \gls{mimoc} running on the \mc in outdoor environments.
\mc was able to traverse different terrains with different slopes and friction including grass, rocks, asphalt, mud, etc. 
The full experiments can be found in~\vref{5}.}
\label{fig_res4}
\end{figure}

In this work, we propose \gls{mimoc}, an~\gls{rl} locomotion controller.
\gls{mimoc} combines the best of both worlds by using model-based optimal control to guide the \gls{rl} policy.
We train our policies to imitate rich reference trajectories 
from model-based optimal controllers~\cite{DiCarlo2018,Kim2019}. 
The \gls{rl} policy trains with simple, minimally-tuned rewards.
To transfer the policy on hardware
as shown in~\fref{fig_res4}, we train on a range of sensor noise, body masses, and disturbances.

\section{Related Work}
In \gls{rl} for legged robots, 
the policy requires some inductive bias or prior knowledge.
One way is to introduce a gait dependency. 
This is usually referred to as~\glspl{pmtg}~\cite{Lee2021, Iscen2018}, 
where the prior is enforced via~\glspl{cpg}, 
and the policy then modulates the parameters of the~\gls{cpg}, such as stepping frequency.
The main drawback of this approach is
the explicit gait dependency; 
the gait is not an emergent behavior but is rather transcribed by the~\gls{cpg}.
Another way to use priors is via reward shaping~\cite{Ji2022, Rudin2022, Siekmann2021}.
Here, the policy does not depend on a specific gait and several gaits may emerge. 
However, the issue is the rewards are handcrafted, may be complicated, and require tedious tuning.

Another common approach is to learn locomotion by imitating motion from animals or 
animal-like characters~\cite{Peng2020, Hasenclever2020, Zhang2018, Peng2018}. 
The prior knowledge is encoded via reference motions the policy is rewarded for tracking.
These reference motions are obtained either from animated motion data, \gls{mocap} data, 
or from video clips of animals and humans.
The reference motions are then retargeted to fit the morphology of the legged robot.

Despite learning different locomotion skills on real legged robots, 
\gls{rl} via motion imitation poses several challenges.
One challenge is in acquiring data, especially \gls{mocap}~data.
It~is especially hard to acquire animal \gls{mocap} data versus human data~\cite{Zhang2018}. 
Additionally, motion retargeting poses another issue
since the morphology of humans and animals are different from humanoid and quadruped robots.
Thus, the retargeted motion may not be kinematically accurate.
Finally, 
there is no guarantee this motion reference data is dynamically feasible on a legged robot~\cite{Peng2020}.

\section{Proposed Approach and Contribution}
\gls{mimoc} is an \gls{rl} locomotion controller that learns from model-based optimal control. 
Unlike other motion imitation \gls{rl} controllers that rely on \gls{mocap} or video clips, 
\gls{mimoc} relies on the reference trajectories provided by model-based controllers.
Thus, \gls{mimoc} does not require any motion retargeting since the reference trajectories are robot-specific. 
Model-based optimal controllers may consider the robot's full whole-body kinematics and dynamics, 
unlike \gls{mocap} or video clips, to create dynamically feasible reference trajectories.
The reference trajectories can come from many~\gls{to}-based planners similar to~\cite{Chignoli2022, Winkler2018b}. 
We can provide \gls{mimoc} with different labeled locomotion skill references including walking and jumping.
This is an advantage over human or animal data 
since from \gls{mocap} data is highly unstructured and difficult to parameterize~\cite{Zhang2018}.

Learning from model-based controllers provides us with torque references which
are not included in \gls{mocap} or video clips. 
This is advantageous for dynamic locomotion where controlling torques and~\glspl{grf} is essential~\cite{Park2017}.
To elaborate, 
the robot must balance itself by generating contact forces through torques 
which kinematic references do not provide~\cite{Park2017, Wieber2006}. 
Torque references give the robot an idea of the forces it should generate when it is in contact.
Since our reference trajectories are dynamically feasible and provide~\gls{mimoc} with torque references, 
transferring the policy to real platforms does not require extensive domain adaptation or reward shaping and tuning. 

\gls{mimoc} not only overcomes the challenges of \gls{rl} via motion imitation 
but also the challenges that arise from model-based controllers. 
Model-based approaches rely heavily on state estimation which is known to have noise issues~\cite{Fahmi2021}. 
We train \gls{mimoc} with sensor noise to make the policy less sensitive to noisy state estimation. 
We also train our agent on a range of environments and robot physical parameters 
to make it more robust than the model-based controller. 

Similar to~\gls{mimoc},  other~\gls{rl}~controllers imitate reference motions 
from model-based controllers~\cite{Babadi2019, Li2021b, Brakel2021}. 
Babadi et al.~\cite{Babadi2019} proposed imitating references from a Monte Carlo-based planner 
and evaluated this approach on bipedal and quadrupedal characters.
Li et al.~\cite{Li2021b} relied on a hybrid-zero-dynamics-based planner to generate the motion references 
and evaluated this approach on the Cassie bipedal robot. 
Brakel et al.~\cite{Brakel2021} proposed a centroidal-dynamics model-based planner to generate the motion references
and tested this approach on the ANYmal quadruped robot. 
\gls{mimoc} differs from the aforementioned works in several aspects.

First, 
the work in~\cite{Babadi2019, Li2021b, Brakel2021} does not include torque reward tracking, 
which we demonstrate as valuable in producing high-quality policies on the real robot.
Second,
the policies in~\cite{Brakel2021, Li2021b} require motion references (future and past motion frames) as observations. 
Thus, these policies are restricted to the reference trajectory behaviors on deployment 
and are unable to generalize beyond that. 
On the other hand, 
\gls{mimoc} uses the reference trajectories only to reward the policy and to initialize each episode; 
not as observations.
As a result, \gls{mimoc} does need a gait library or a reference generator when deployed.
Additionally, \gls{mimoc} shows that motion references are not needed as observations, 
which reduces the input size of the policy's network.

Third, in this work, the reference trajectories do not come directly from model-based optimization. 
Instead, we run our model-based controllers in a highly accurate dynamic simulator and use the resulting 
motion of the robot as the reference. 
This provides us highly accurate, dynamically feasible references.
Finally,
the work in~\cite{Brakel2021} was tested only on a single robot in simulation, 
the work in~\cite{Babadi2019} was not tested on hardware, 
and the work in~\cite{Li2021b} was tested only on a bipedal robot. 
In this work,
we test \gls{mimoc} on the \humanoid~\cite{Chignoli2021} 
and the \mc~\cite{Katz2019} in two different simulation environments: 
\ig~\cite{Makoviychuk2021}, and \RS~\cite{RobotSoftware}. 
We also test \gls{mimoc} on the \mc hardware.

\section{Method}\label{sec_methods}
\begin{figure}[t]\centering
\includegraphics[width=\columnwidth]{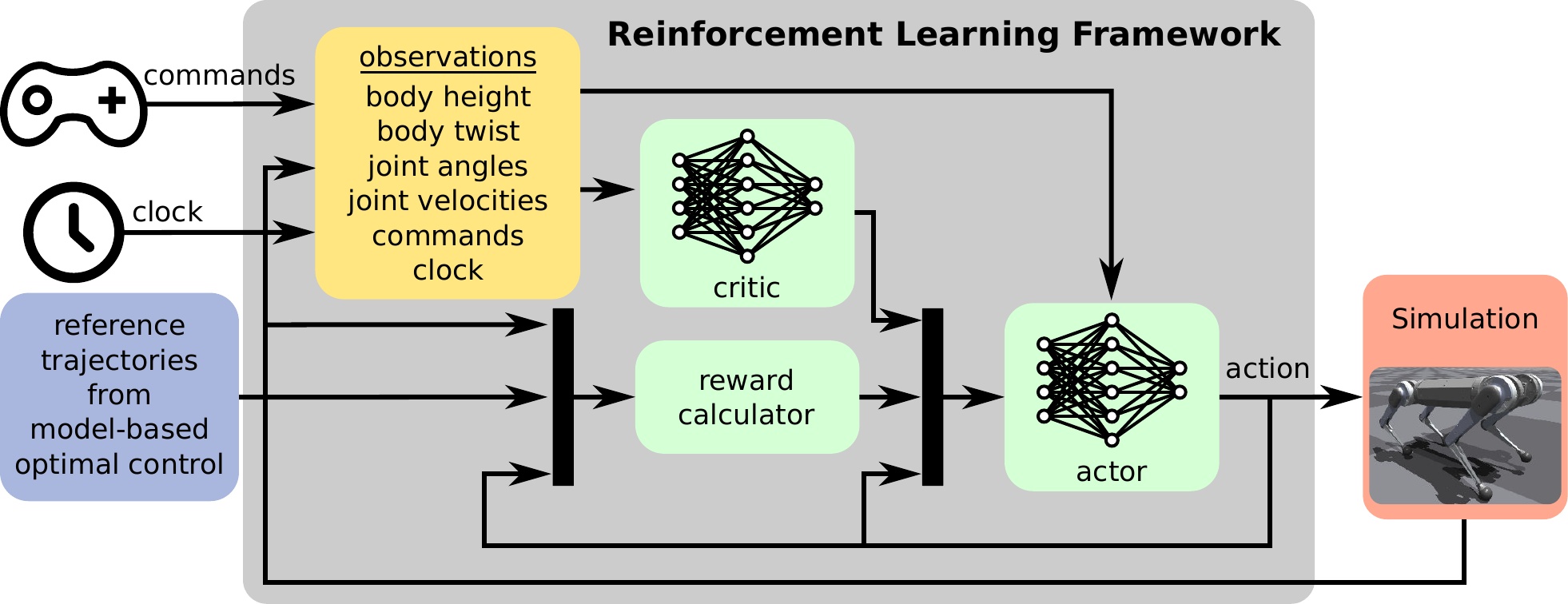}	
\caption{The~\gls{rl} framework used to train~\gls{mimoc}'s policy.}
\label{fig_1}
\end{figure}

The goal of \gls{mimoc} is to learn a locomotion controller that tracks user commands by imitating model-based optimal control reference trajectories.
An overview of the~\gls{rl} framework used during training is shown in~\fref{fig_1}. 
The reference trajectories are only used for reward computation and episode initialization for training.
The policy takes the observed robot states and commands as input
and outputs joint position set points that are sent to the joint-level feedback controller.
Unlike~\cite{Peng2020} and~\cite{Brakel2021}, respectively, 
\gls{mimoc} is trained without motion retargeting and domain adaptation and also without fine-tuning.

\subsection{Model-Based Optimal Control Reference Trajectories}\label{sec_ref_traj}
The reference trajectories for the \mc and the \humanoid 
are generated from the model-based framework in~\cite{Kim2019}. 
It includes an~\gls{mpc} and a~\gls{wbc} 
running hierarchically at different update rates.
We collect the reference trajectories by simulating the controllers in the \RS simulator 
given a range of scripted commands and using the ground truth robot states. 
We use \RS~\cite{RobotSoftware} because it accounts for the dynamic effects of the robot’s motors 
and is thus more accurate than generic simulators.

To generate the reference trajectories, 
we collect the robot's \text{state~$s\in\Rnum^{3n+12+2n_e}$}, the phase clock~$\phi\in\Rnum$, 
and the commands $c\in\Rnum^3$ at every time step~$t$.
The robot state is defined as
\begin{equation}
s = [q^T \quad \dot{q}^T  \quad \tau^T \quad 
x^T \quad \dot{x}^T \quad \theta^T \quad \omega^T \quad e^T \quad \dot{e}^T]^T
\end{equation}
where~$q\in\Rnum^n$, $\dot{q}\in\Rnum^n$, 
and $\tau\in\Rnum^n$ are the vectors of joint positions, velocities, and torques, respectively, 
and~$n$ is the number of joints. 
The vectors~$x\in\Rnum^3$, $\dot{x}\in\Rnum^3$, $\theta\in\Rnum^3$, and $\omega\in\Rnum^3$
are the body (floating base) positions, linear velocities, orientation in Cardan angles (roll-pitch-yaw), 
and angular velocities, respectively. 
The vectors $e\in\Rnum^{n_e}$, $\dot{e}\in\Rnum^{n_e}$ are the end-effector positions and velocities, respectively,
and $n_e$ is the number of end-effector degrees of freedom 
(number of end-effectors multiplied by the number of degrees of freedom per end effector). 
Finally, the commands~$c = [c_{vx}, \ \ c_{vy}, \ \ c_{\dot{\psi}}] \in\Rnum^3$ 
are the body forward velocity, lateral velocity, and yaw rate, respectively. 

The commands are scripted to move the robot in this order:
forward then backward with a longitudinal velocity of~$\pm \bar{c}_{vx}$, 
leftwards then rightwards with a lateral velocity of~$\pm \bar{c}_{vy}$, 
anti-clockwise then clockwise with a yaw rate of~$\pm \bar{c}_{\dot{\psi}}$.
The values of~$\bar{c} = [\bar{c}_{vx}, \ \bar{c}_{vy}, \ \bar{c}_{\dot{\psi}}]$ 
are $[0.5, 0.2, 2.0]$ for the \mc and the \humanoid.
The final dataset consisted of a single clip of \unit[60]{s} for the \mc and \unit[180]{s} for the \humanoid. 
Videos showing the generated reference trajectories for the \mc and the \humanoid
are shown in \vref{6} and \vref{7}, respectively.

\subsection{\gls{mimoc}'s \gls{rl} Framework}\label{sec_rl}
\gls{mimoc} is formulated as an \gls{rl} control problem~\cite{Sutton2018}.
The \gls{rl} control policy~$\pi(a\vert o)$ is the mapping between observations~$o$ and actions~$a$.
At~each time step~$t$, 
given the robot's current observation~$o_t$ 
the agent samples an action~$a_t$ from the policy~$\pi(a_t\vert o_t)$ 
and receives a reward $r_t = r(s_t, a_t)$ accordingly.
When the action is applied to the agent, the environment emits the next observation~$o_{t+1}$.
The goal of~\gls{rl} is to find the optimal policy that maximizes the expected reward at each time step.  
We formulate our \gls{rl} problem as an \gls{ac} method and solve it using \gls{ppo}~\cite{Schulman2017}.

\boldSubSec{Gym Environment}
We used \ig~\cite{Makoviychuk2021} with the Legged Gym implementation~\cite{Rudin2022}.
The policy is trained with 4096 simultaneous agents on flat terrain.
The simulation environment runs at \unit[500]{Hz}, and the action is computed every \unit[5]{time steps}.
Thus, the policy runs at \unit[100]{Hz}.

\boldSubSec{Episode}
At the beginning of an episode, 
the agents are initialized with a state that is randomly sampled from the reference trajectory.
Instead of running the episode until the end of the trajectory, 
we run the episode for a length varying between \unit[1-5]{s} depending on the training task.

\boldSubSec{Termination}
Episodes are terminated and restarted under three conditions.
First, if the length of the episode exceeds the maximum episode length.
Second, if the end of the reference trajectory is reached.
Third, if any robot link other than the feet collides with the ground (\ie ground collision).

\boldSubSec{Rewards}
The reward function is designed to guide the policy to track the reference trajectories during training. 
To track a reference trajectory is to minimize the error (deviation) between a certain robot state variable~$\beta$
and it's corresponding desired reference~$\beta_{\mathrm{ref}}$ (\ie minimize $\beta_{\mathrm{ref}} - \beta$).
To minimize this error, we design the reward to maximize the squared exponential of this error 
(\ie maximize $\mathrm{exp}(-\|\beta_{\mathrm{ref}} - \beta\|^2)$).
Since, the reference trajectories include the body, joint, and end-effector positions and velocities, and joint torques,
we exploit all of these references in the reward function.
This reward function is inspired from \cite{Peng2020, Hasenclever2020}. The accumulated reward is computed~as
\begin{equation}
r_t = r_{q} + r_{\dot{q}} + r_{\tau} + r_{x}+ r_{\dot{x}} + r_{\theta} + r_{\omega} + r_{e} + r_{\dot{e}}
\label{rewards}
\end{equation}
where each term maximizes the squared exponential of the error of a certain robot state variable~$\beta$. 
The terms in~\eref{rewards} correspond to 
the robot joint position~$r_{q}$, velocity~$r_{\dot{q}}$, and torques~$r_{\tau}$, 
body position~$r_x$, linear velocity~$ r_{\dot{x}}$, orientation~$r_{\theta}$, and angular velocity~$r_{\omega}$, 
end-effector positions~$r_{e}$ and velocities~$r_{\dot{e}}$. 
Each term in the reward function is of the form of the generic reward $r_{\beta}$
\begin{equation}
r_{\beta} = w_{\beta} \sum_{i=1}^{n_{d}}  \Bigg[ \mathrm{exp}( - \alpha_{\beta}   \big\|  \beta_{i,\mathrm{ref}} - \beta_i  \big\|^2 ) \Bigg]
\label{rewardss2}
\end{equation}
which is a weighted squared exponential function of weight~$w_\beta$ and width~$\alpha_\beta$.
The weights~$w_\beta$ are manually tuned but the widths~$\alpha_\beta$ are computed as
\begin{equation}
\alpha_\beta = \left(\frac{2}{\max(\beta^{\mathrm{ref}}) -\min(\beta^{\mathrm{ref}}) }\right)^2.
\label{eq_alpha}
\end{equation}
The values of the  weights~$w_\beta$ and widths~$\alpha_\beta$
for \mc and the \humanoid are in~\tref{tab_rew_mc}.
We found that unit weights~($w_\beta$=1) were sufficient to train the policies.
However, we found that increasing the weights~$w_\beta$ of the joint positions~$w_q$ and torques~$w_\tau$ helped in reducing the training time of the \humanoid policy.
More importantly, we found that tuning the widths of the squared exponential rewards~$\alpha_\beta$ was crucial. 
Without tuning these widths based on~\eref{eq_alpha}, the policies would fail to track the references.

\boldSubSec{Observations}
The observations~$o_t\in\Rnum^{2n+1+6+3+2}$ are defined as
\begin{equation}
o_t = [q^T \quad \dot{q}^T \quad 
z^T \quad \dot{x}^T \quad \omega^T \quad 
c^T \quad \cos(\phi) \quad \sin(\phi)]^T
\end{equation}
where $z\in\Rnum$ is the body height, and $\cos(\phi)$ and $\sin(\phi)$ are associated with the gait phase clock~$\phi$.
The commands~$c\in\Rnum^3$ come from the reference trajectories during training 
but are user-input during deployment.
The body's linear velocity is expressed in body-fixed frame.
Finally, these observations were not normalized or scaled.

\boldSubSec{Actions and Control Law}
The actions~$a_t\in\Rnum^{n}$ are defined as joint position residuals 
from default joint positions $q_0$ (\ie the robot's default standing configuration). 
Thus, the desired torques~$\tau_d$ sent to the robots are computed~as
\begin{equation}
\tau_{d} = k_p(k_a a_t + q_0 - q_t) - k_d \dot{q}_t
\label{torques_rl}
\end{equation}
where $k_p$ and $k_d$ are the proportional and derivative gains, respectively, and~$k_a$ is the action gain. 
The gains used for all of the joints of the \mc are: $k_a=0.25$, $k_p = 20$, and $k_d=0.5$.
Each side of the \humanoid (9~joints per side - 4~arm joints and 5~leg joints) had the following gains:
$k_a = 0.5$, 
$k_p = [60, 60, 60, 60, 40, 10, 10, 10, 10]$, and
$k_d = [5, 5, 5, 5, 0.1, 0.5, 0.5, 0.5, 0.5]$.

\boldSubSec{Network Architecture}
The actor and critic networks are defined as~\gls{mlp} networks with ELU activations and
hidden dimensions of $256\times256\times256$.

\boldSubSec{Domain Randomization}
We added noise to the observations with random values with a range of~$[-1, 1]$ 
multiplied by a scale for each observation.
We also randomized the terrain friction with a range of~$[0.5, 1.25]$ and
the robot mass with a range of~$[-2, 2]$~\unit{kg}. 
We added disturbance to the robot's body 
by applying an instant change of the body linear velocity with a range of~$[-1, 1]$~\unit{m/s}.

\boldSubSec{Policy Training}\label{sec_hyper_param}
The policies are trained using the \gls{ppo} implementation in \cite{Rudin2022}.
\mc is trained for as few as \unit[500~]{iterations} in under \unit[10~]{minutes} of training (on an Intel 12900K CPU and Nvidia 3080ti GPU), 
but we preferred training for \unit[1500~]{iterations}.
\humanoid is trained with iterations ranging between \unit[1500~]{iterations} and \unit[10000~]{iterations}.
We used a
learning rate of $10^{-3}$,
batch size of 98304,
24 steps per policy update, 
mini-batches of 4, 
value loss of 1.0,
clip parameter of 0.2, 
entropy coefficient of 0.01, 
discount factor of 0.99, 
GAE discount factor of 0.95, and 
desired KL of 0.01.

\boldSubSec{Real World Deployment}\label{sec_dep}
The policy is evaluated on board the robot at~\unit[100]{Hz}.
To compute the observations, we relied on a contact-based state estimator
similar to the state estimator in~\cite{Bledt2018}.
The actions are sent to a low-level joint controller that runs at~\unit[40]{kHz}.

\begin{table}[t!]\label{tab_rew_mc}
\centering \caption{Reward Scales of the \mc and \humanoid.}
\renewcommand{\arraystretch}{1.2}
\begin{tabular}{cl}
\hline \hline
Weight~$w_\beta$ & Width~$\alpha_\beta$      \\
\hline Cheetah & ~\\
$w_q = 1.0$           &  $\alpha_q = [0.1, 0.5, 0.5]$         $\in \Rnum^3$ per leg\\    
$w_{\dot{q}} = 1.0$   &  $\alpha_{\dot{q}} = [15, 20, 25]$    $\in \Rnum^3$ per leg\\        
$w_\tau = 1.0$        &  $\alpha_\tau = [5.0, 5.0, 8.0]$      $\in \Rnum^3$ per leg\\   
$w_x = 1.0$           &  $\alpha_x = [0.02, 0.02, 0.02] $     $\in \Rnum^3$\\          
$w_{\dot{x}} = 1.0$   &  $\alpha_{\dot{x}} =[0.6, 0.1, 0.5]$  $ \in \Rnum^3$\\          
$w_\theta = 1.0$      &  $\alpha_\theta = [0.31, 0.31, 0.31]$ $\in \Rnum^3$\\         
$w_\omega = 1.0$      &  $\alpha_\omega =[3.3, 2.5, 3.7] $    $ \in \Rnum^3$\\         
$w_e = 1.0$           &  $\alpha_e =  [0.02, 0.02, 0.02]$     $\in \Rnum^3$ per foot\\               
$w_{\dot{e}} = 1.0$   &  $\alpha_{\dot{e}} = [0.6, 0.1, 0.5]$ $\in \Rnum^3$ per foot\\   
\hline Humanoid & ~\\
$w_q = 5.0$           &  $\alpha_q = [0.2, 0.3, 0.9, 0.9, 0.5, 0.2, 0.2, 0.2, 0.5]$ $\in \Rnum^9$\\    
$w_{\dot{q}} = 1.0$   &  $\alpha_{\dot{q}} = [3,5,10,12,8,8,1,1,12]$  $\in \Rnum^9$ per side\\       
$w_\tau = 10.0$       &  $\alpha_\tau = [10,40,50,60,20,1,1,1,1]$     $\in \Rnum^9$  per side\\   %
$w_x = 1.0$           &  $\alpha_x = [0.06 ,0.06,0.06] $     $\in \Rnum^3$\\          
$w_{\dot{x}} = 1.0$   &  $\alpha_{\dot{x}} = [1.03,0.59,0.61]$  $\in \Rnum^3$\\ %
$w_\theta = 1.0$      &  $\alpha_\theta = [0.27,0.29,0.13]$ $\in \Rnum^3$\\         
$w_\omega = 1.0$      &  $\alpha_\omega = [4.01,2.98,2.04] $    $\in \Rnum^3$\\         %
$w_e = 1.0$           &  $\alpha_e =  [2, 0.1,0.1, 2, 0.1, 0.1]$     $\in \Rnum^6$ per side\\ 
$w_{\dot{e}} = 1.0$   &  $\alpha_{\dot{e}} = [2, 1, 2, 2,1,2]$ $\in \Rnum^6$ per side\\       
\hline \hline
\end{tabular}
\end{table}

\section{Results}
We evaluate \gls{mimoc} 
on \mc (in simulation and experiment) and the \humanoid (in simulation), 
and we compare \gls{mimoc} with the~\gls{cmpc} model-based controller detailed in~\cite{Kim2019, DiCarlo2018}. 
To test our policies and compare them with the model-based controller, we used \RS~\cite{RobotSoftware}
because it is a more realistic simulation environment.
Thus, the simulation would be one step toward real-world deployment, 
and we can evaluate how well can the policy be transferred to a different simulator.
The videos supporting the results can be found in~\cite{Video}.

\begin{figure}[t!]\centering
\includegraphics[width=0.9\columnwidth]{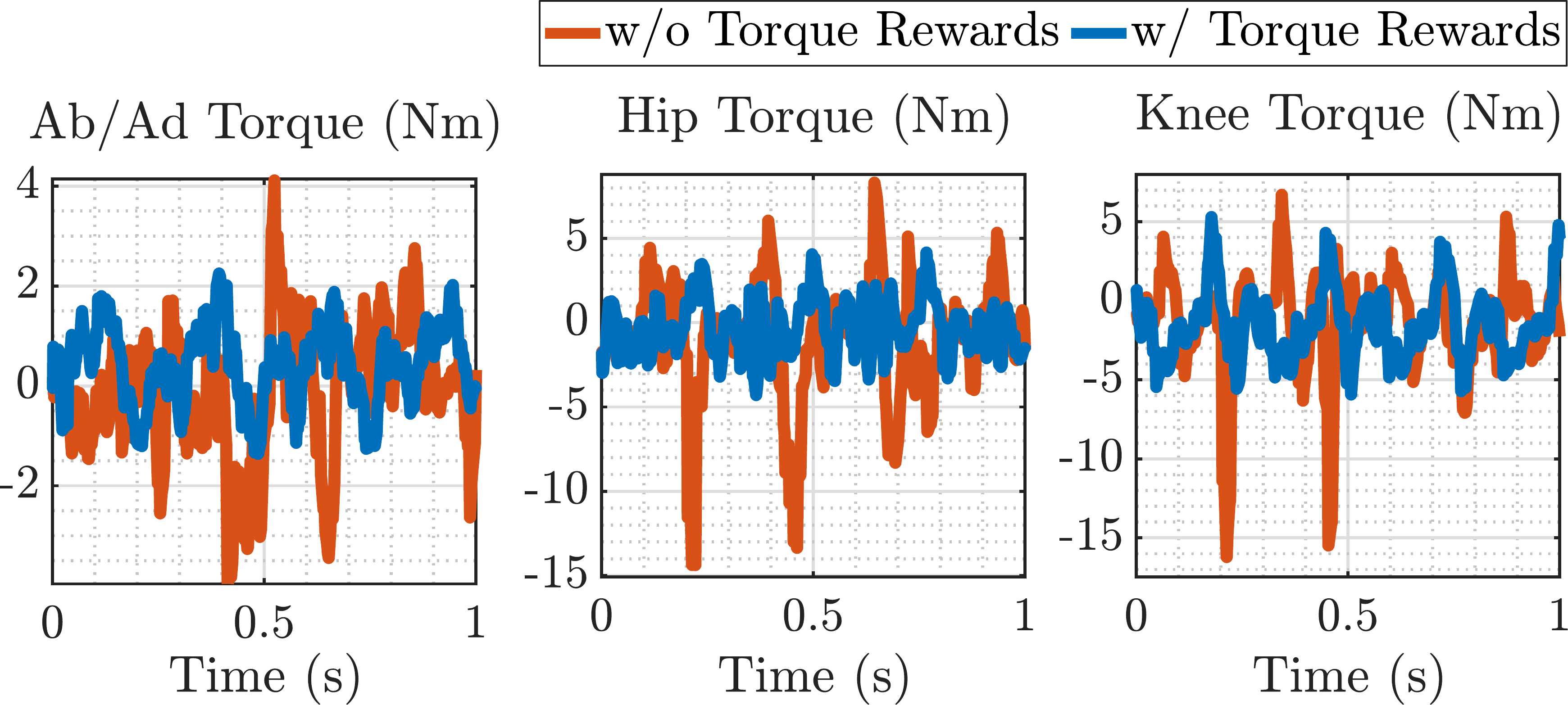}		
\caption{The Importance of Torque Tracking Rewards.
Two policies tested on the \mc: with versus without torque tracking rewards.
The figure shows the torque values of the three joints of the right-front leg of the \mc in an experiment.}
\label{fig_res1}
\end{figure}

\begin{figure}[t!]\centering
\includegraphics[width=0.9\columnwidth]{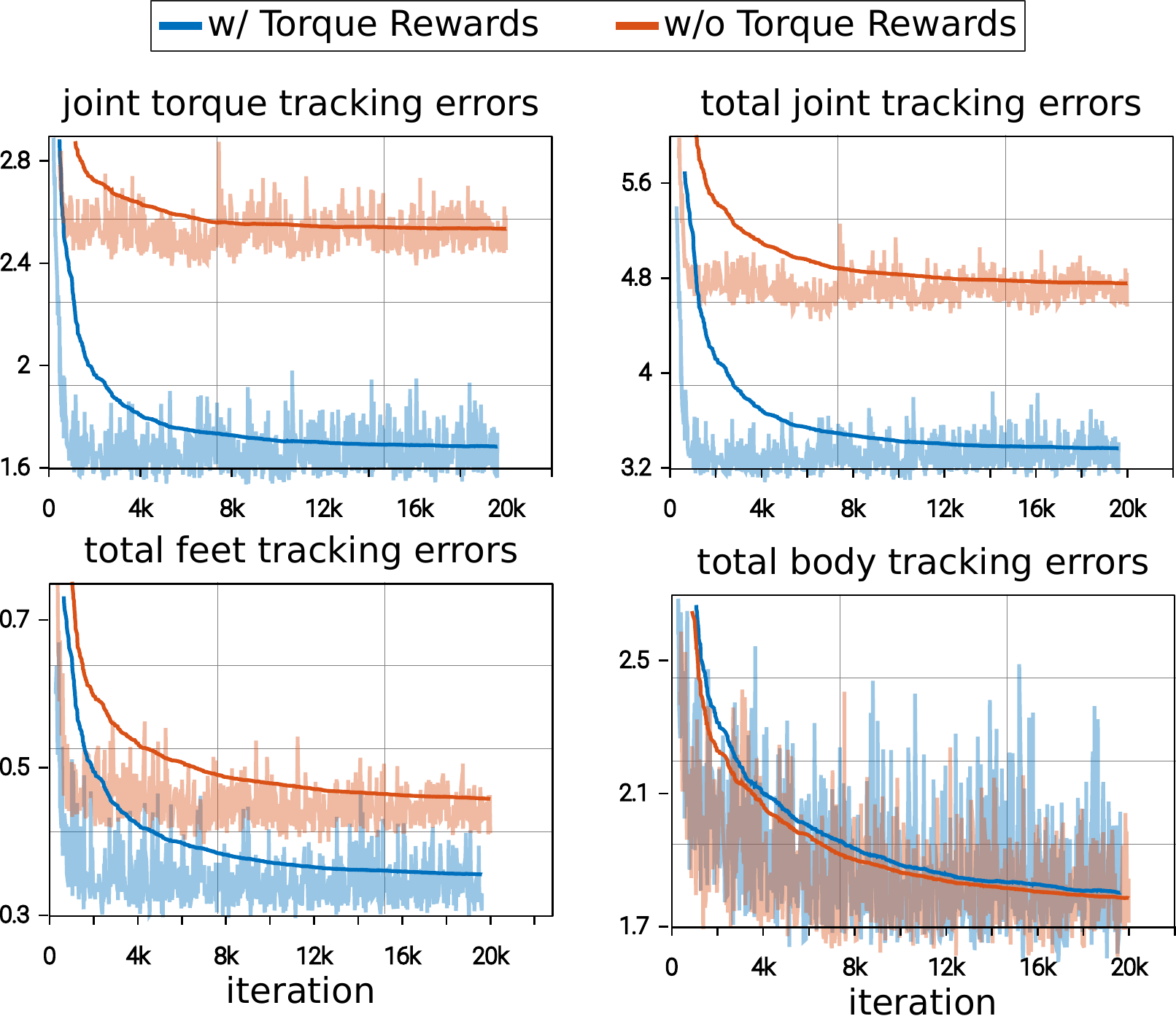}		
\caption{The Importance of Torque Tracking Rewards.
Two policies tested on the \mc: with versus without torque tracking rewards.
The figure shows the tracking errors of the torques, and the total tracking errors of the joints, feet and body.
The tracking error is defined as the mean over all agents of
the difference between the robot's current state and its corresponding reference.
}
\label{fig_sup_res0}
\end{figure}
\boldSubSec{The Importance of Torque Tracking Rewards}
\gls{mimoc}'s primary advantage over motion imitation from \gls{mocap} 
is that our model-based reference trajectories include torque references.
We hypothesize imitating reference torques is essential for learning
because controlling torques and \glspl{grf} are as important as controlling positions and velocities; especially during dynamic locomotion. 
To verify this, we compared the performance of two \gls{mimoc} policies: one trained with and another without torque tracking reward.
We ran these policies on the \mc as shown in~\vref{1}.
Figure~\ref{fig_res1} shows the torque values of the three joints of one leg of the \mc.
As shown in the video, the policy trained without torque tracking rewards was shaking substantially more on the hardware. 
Without the torque tracking rewards, the feet had a more aggressive contact with the ground.
This is also evident in~\fref{fig_res1}, where the no torque tracking reward policy had higher torque peaks.
Additionally, as shown in~\fref{fig_sup_res0},
when~\gls{mimoc} is trained with torque tracking rewards, the policy learns faster with smaller tracking errors (deviations from the reference trajectories).

\begin{figure}[t!]\centering
\includegraphics[width=\columnwidth]{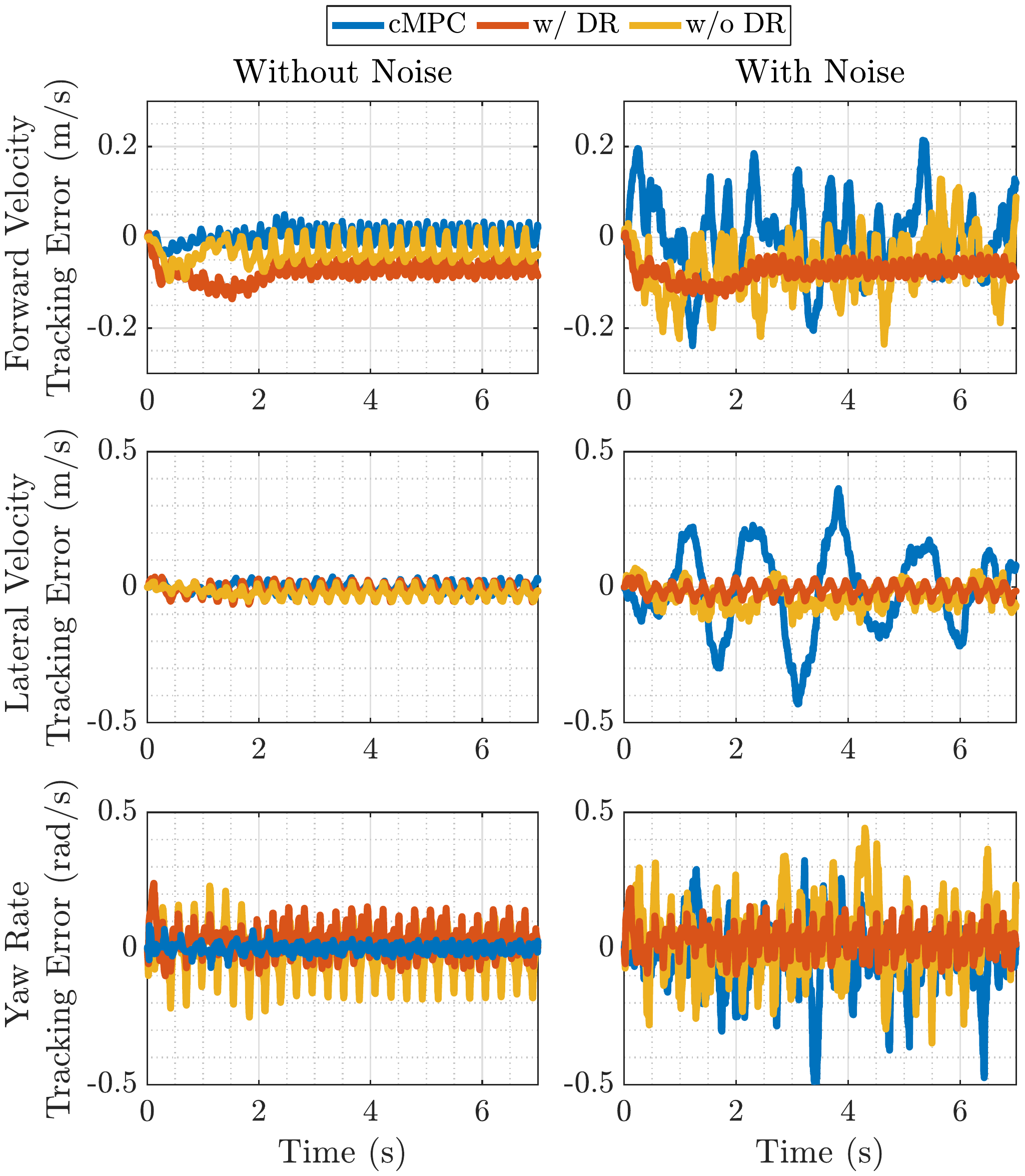}		
\caption{Performance under Noisy State Estimation. 
The controllers evaluated were \gls{mimoc} with domain randomization, \gls{mimoc} without domain randomization, and model-based~\gls{cmpc}. 
Left: the tracking errors of the three controllers with perfect (noise-free) state estimates.
Right: the tracking errors of the three controllers with noisy state estimates.}
\label{fig_res2}
\end{figure}

\boldSubSec{Performance under Noisy State Estimation}
A key advantage of \gls{rl} over model-based 
is that~\gls{rl} may be less sensitive to noise and/or model uncertainties, 
because the policy can be trained with noisy observations as part of domain randomization.
To verify this, 
we evaluated \gls{mimoc} with domain randomization, 
\gls{mimoc} without domain randomization, 
and~\gls{cmpc} in simulation (\RS) under noisy state estimates.
To simulate noisy state estimates, 
we inject noise onto the robot's body states (linear and angular position and velocity)
following a normal distribution $X$ with mean $\mu$ and standard deviation $\sigma$: $X \sim \mathcal{N}(\mu,\,\sigma^{2})$.
We used a mean $\mu=0$ for all body states 
and standard deviations $\sigma$~of $0.03$,~$0.15$,~$0.075$, and~$0.04$ 
for the position, orientation, linear velocity, and angular velocity of the robot's body, respectively. 
We commanded \mc to trot forward with a velocity of \unit[0.5]{m/s}, 
once with ground truth states (without noise) and once with noisy state estimates.

Figure~\ref{fig_res2} shows the tracking error of the forward and lateral velocity and the yaw rate from the two~\gls{mimoc} policies and from~\gls{cmpc}.
As shown in~\fref{fig_res2}, with perfect state estimation
the three controllers have small tracking errors with slightly less error for the model-based controller and slightly more tracking error for~\gls{mimoc} without domain randomization. 
However, if the state estimates are noisy,
the model-based controller performs the worst followed by~\gls{mimoc} without domain randomization.  
This demonstrates the importance of domain randomization during training 
and that~\gls{mimoc} may outperform~\gls{cmpc} in the real-world where state estimation is indeed noisy.
As shown in~\vref{2}, \mc did not fall from any of the controllers. 
Yet, with noisy state estimates, the model-based controller was less stable than~\gls{mimoc}.

\begin{figure}\centering
\includegraphics[width=\columnwidth]{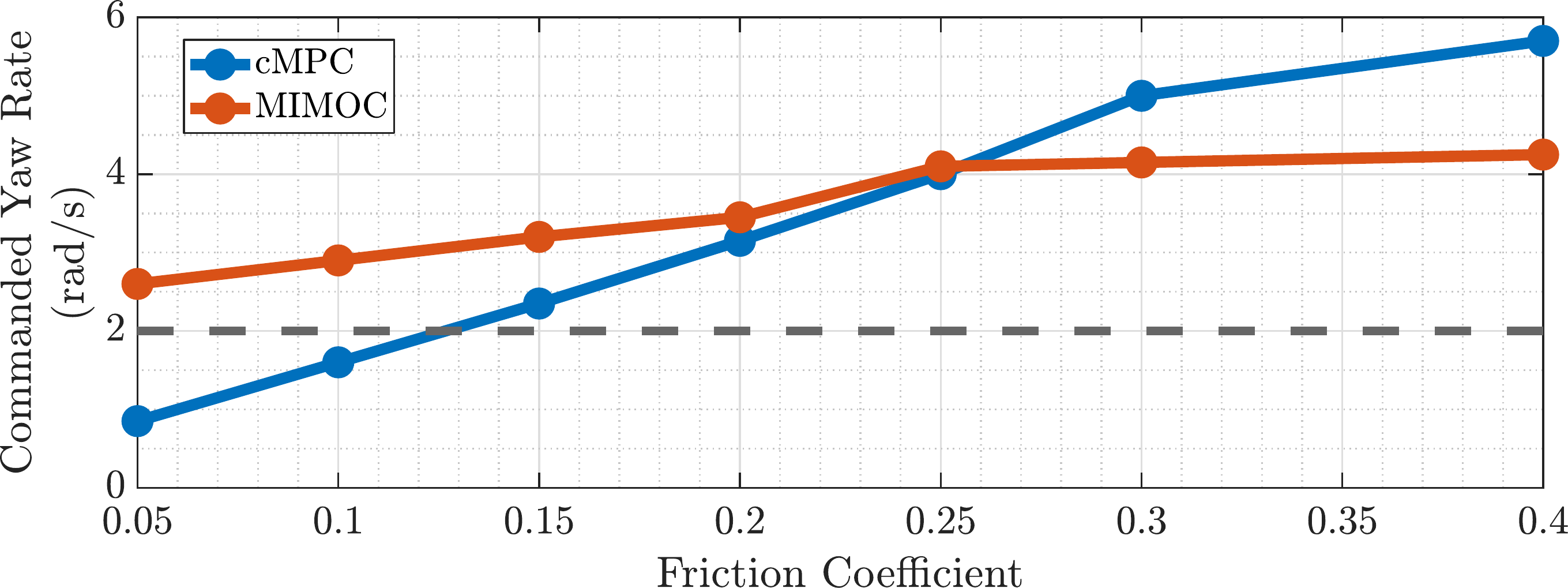}		
\caption{Performance under Different Terrain Friction.
The figure shows the maximum yaw rate command that \mc was able to withstand at a given friction coefficient, using~\gls{mimoc} and~\gls{cmpc}. The dashed gray line is the maximum yaw rate that~\gls{mimoc} was trained with.}
\label{fig_res3}
\end{figure}

\boldSubSec{Performance under Different Terrain Friction}
To compare \gls{mimoc} against \gls{cmpc} under different terrain friction coefficients,
we evaluate the performance of \mc with different yaw rate commands and friction coefficients in simulation (\RS). 
We simulated 8 different friction coefficients ranging from 0.05~to~0.4 as shown in~\fref{fig_res3}. 
For every friction coefficient, we command \mc to follow a certain yaw rate.
Then, we increment the yaw rate command until the robot falls. 
For all friction coefficients,~\gls{mimoc} was able to withstand higher yaw rate commands than it was trained with;
it trained with a maximum yaw rate of~\unit[2.0]{rad/s}.
\gls{mimoc} outperformed~\gls{cmpc} for friction coefficients below~\unit[0.25]
but \gls{cmpc} outperformed~\gls{mimoc} for friction coefficients above~0.25.
\vref{3}~shows \gls{mimoc} and~\gls{cmpc} in these simulations.

\begin{figure}[t!]\centering
\includegraphics[width=\columnwidth]{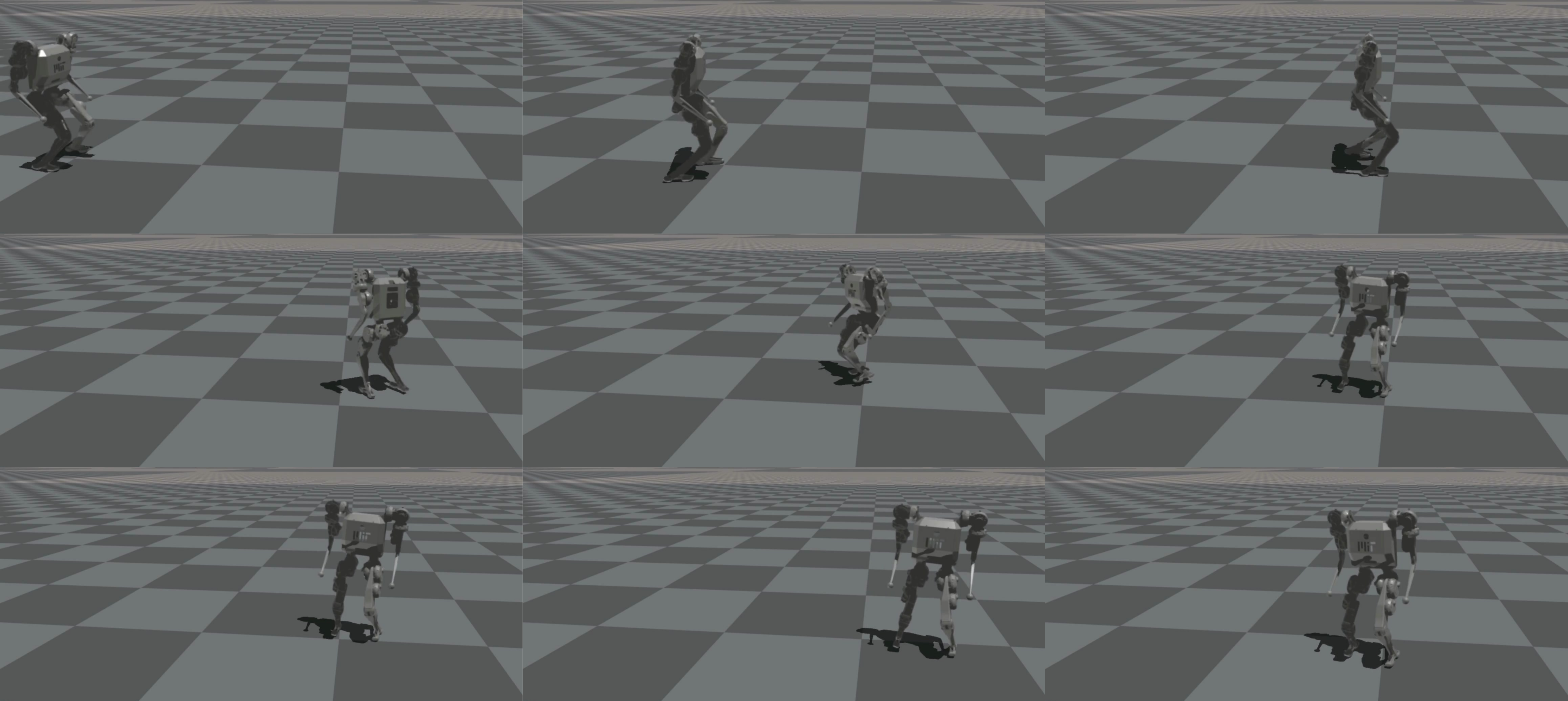}		
\caption{This figure shows multiple screenshots of \gls{mimoc} on the \humanoid. 
We can command the \humanoid to walk in all directions separately and simultaneously, 
something not included in the reference trajectories and \gls{mimoc} was not trained for.}
\label{fig_res_humanoid}
\end{figure}

\boldSubSec{\gls{mimoc} on the \humanoid}
To show that \gls{mimoc} can be applied to other legged platforms, we tested the~\humanoid with~\gls{mimoc} in \ig.
In the first simulation, we ran the policy on the \humanoid and commanded it to walk in multiple directions. 
Figure~\ref{fig_res_humanoid} shows screenshots of the \humanoid walking in \ig and~\vref{4} shows the full simulation.
As shown in \fref{fig_res_humanoid} and \vref{4}, 
we could command the \humanoid to walk in all directions separately and simultaneously. 
Note that the reference trajectories did not include
instances where the robot was commanded in multiple directions. 
The scripted commands were designed 
to direct the robot to walk in one direction at a time.
This shows us that \gls{mimoc} can generalize beyond the reference trajectories. 

\begin{figure}[t!]\centering
\includegraphics[width=\columnwidth]{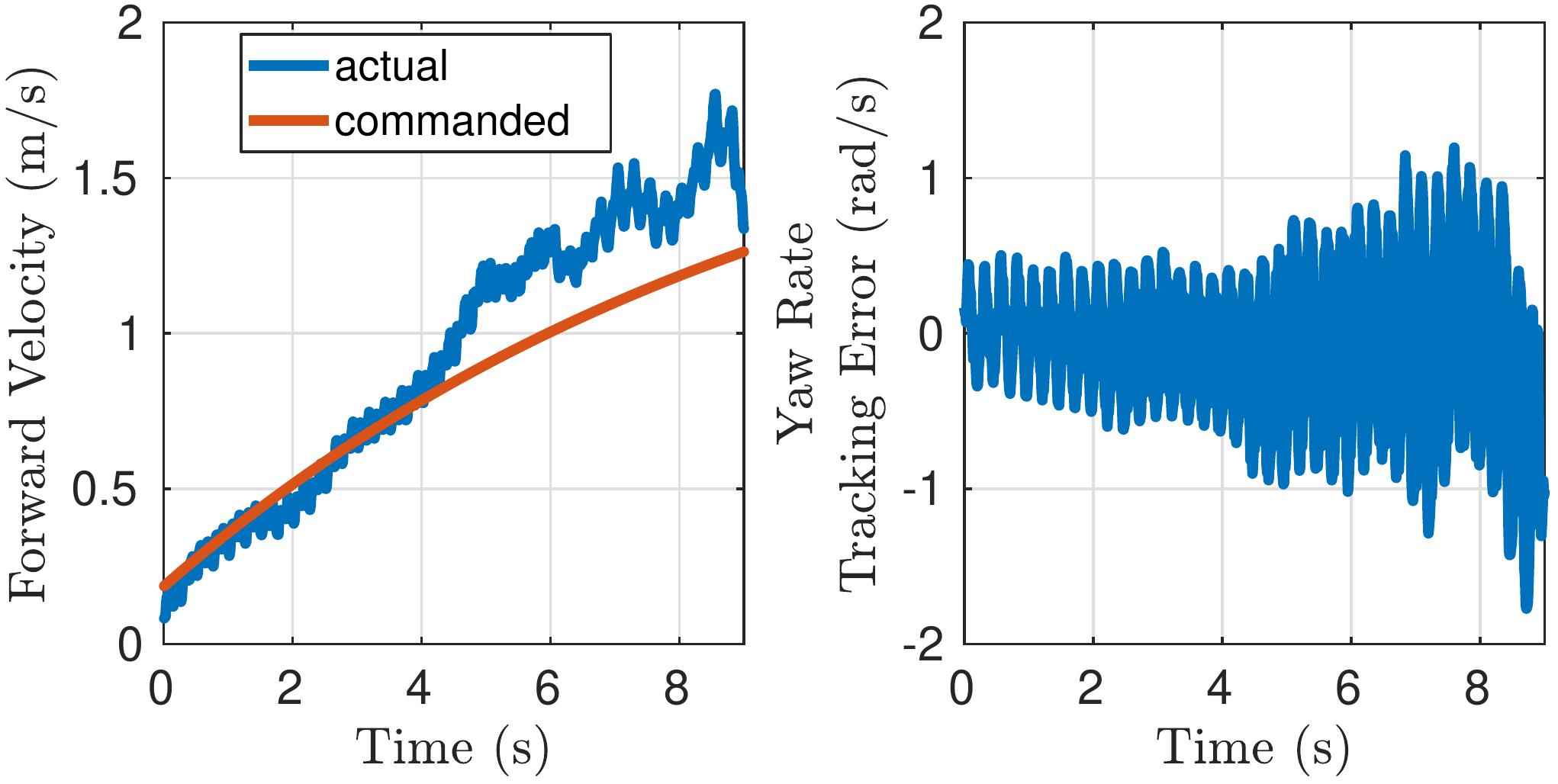}		
\caption{The actual and commanded forward velocity (left) and the tracking error of the yaw rate (right).
The figure shows the tracking performance of the \humanoid walking using \gls{mimoc} and how fast it can walk.}
\label{fig_res_humanoid_tracking}
\end{figure}

In another simulation, we evaluate how fast can the \humanoid walk using \gls{mimoc}. 
To do so, we created a simulation where we command the robot to move forward
with increasing forward velocity. 
The simulation starts with the robot stepping in-place
and ends when the robot falls.
Figure~\ref{fig_res_humanoid_tracking} shows the actual and forward velocity, and
the tracking error of the yaw rate. 
As shown in \fref{fig_res_humanoid_tracking}, 
the robot was able to walk 
with a forward velocity exceeding~\unit[1]{m/s}.
Over~\unit[1]{m/s}, the robot was still able to walk, but the tracking errors
increased. 
In this simulation, the robot reached a maximum commanded forward velocity of~\unit[1.26]{m/s}.

\boldSubSec{\mc Outdoor Experiments}
To evaluate the robustness of~\gls{mimoc}, 
we tested it on \mc in several outdoor environments. 
Figure~\ref{fig_res4} shows screenshots of these experiments shown in more detail in~\vref{5}.
\mc with \gls{mimoc} traversed multiple terrains following user commands.
We could also command \mc at different gait frequencies, something that~\gls{mimoc} was not trained for.
We also commanded \mc for higher velocities than in training. 
As shown in~\vref{5}, another advantage of \gls{rl} over~\gls{cmpc} is~\gls{mimoc} does not go unstable if we suddenly lift \mc.
If~\gls{cmpc} were lifted, the robot would suddenly go unstable and flail wildly.
Finally, as shown in~\vref{5}, \mc was able to traverse different terrains
with different slopes and friction including grass, rocks, asphalt, mud, etc.

\section{Limitations and Future Work}
We identified several limitations of \gls{mimoc} that we are investigating for future work.
First, although we randomized body pushes during training, 
we found that the recovery of the \gls{rl}~policy is only reactive.
By this we mean that while the robot can reject some pushes, it is not able to reason about taking steps
if it deviates substantially from the trained reference.
Second, the references only include trajectories from locomotion over flat terrain, 
and the policy was also only trained on flat terrain. 
When we deployed \gls{mimoc} on the \mc in outdoor environments (see \fref{fig_res4}) 
the robot was able to walk over more challenging terrains, 
but we would like to explore better training for more task diversity.
Third, our references only included commanding the robot in a single direction at a time. 
In simulation and experiment, 
\gls{mimoc} was able to generalize to higher speeds and in different directions simultaneously, 
but a more comprehensive dataset would likely improve performance.

To expand on this work and overcome these limitations, we are exploring two main ideas.
One is to use~\gls{mimoc} as a warm-started actor for more traditional reward-shaping and domain randomization.
A second is to use our method to
blend and fuse multiple references into a single policy, 
either by splicing incompatible references or by training a single policy to imitate the behaviors of multiple policies.
We additionally plan to test \gls{mimoc} on the \humanoid hardware.

\section{Conclusion}
We presented \gls{mimoc}, an \gls{rl} locomotion controller that learns by imitating model-based optimal controllers. 
\gls{mimoc} is trained to track reference trajectories provided by a model-based controller and follow user commands.
\gls{mimoc} addresses some challenges faced by other motion imitation \gls{rl} approaches.
\gls{mimoc} does not require motion retargeting since the reference trajectories are designed for the robot, are dynamically feasible because they were captured in dynamic simulation,
and include torque data which we showed to improve training and  performance.
\gls{mimoc} also overcomes some challenges faced by model-based optimal controllers
since it is less sensitive to noisy state estimation because the policy is trained with domain randomization and noisy observations.
Finally, \gls{mimoc} can be deployed on real robots like the \mc
and transfers to other legged systems like the \humanoid.
We showed the robustness of~\gls{mimoc} on the \mc in a wide range of challenging terrain 
including slopes, grass, asphalt, mud, etc. 

\gls{mimoc} has three main outcomes. 
First, torque tracking is indeed essential for real-world deployment and faster learning convergence.
Second, \gls{rl} policies trained with \gls{mimoc} can outperform model-based optimal controllers when state estimation is noisy. 
Third, \gls{mimoc} trained with domain randomization can outperform model-based optimal controllers at low ground friction.

\bibliographystyle{IEEEtran} \bibliography{./includes/bibliography.bib}
\end{document}